\documentclass[11pt]{article}

\usepackage[margin=1in]{geometry}
\usepackage[T1]{fontenc}
\usepackage[utf8]{inputenc}
\usepackage{lmodern}
\usepackage{microtype}
\usepackage{amsmath}
\usepackage{amssymb}
\usepackage{graphicx}
\usepackage{booktabs}
\usepackage{float}
\usepackage{subcaption}
\usepackage{authblk}
\usepackage[hidelinks]{hyperref}
\usepackage{url}

\title{Single Image Super-Resolution via Semantic Prompt and Cross-Modal Training on Content Area}

\author{Xiao Tan}

\affil{Xidian University, Xi'an 710071, China}

\date{}

\hypersetup{
  pdftitle={Single Image Super-Resolution via Semantic Prompt and Cross-Modal Training on Content Area},
  pdfauthor={Xiao Tan},
  pdfkeywords={Image Super-Resolution, Multimodality, Computer Vision, Deep Learning}
}

\begin{document}

\maketitle

\begin{center}
\small
E-mail address: 21171213976@stu.xidian.edu.cn
\end{center}

\begin{abstract}
Enhancing image quality through super-resolution is crucial in image and video processing. While contemporary deep learning models, such as CNNs and Transformers, excel in pixel-level enhancements, they often overlook the rich semantic context within images. To address this, we introduce a novel approach that synergizes pixel intensity with semantic information for superior detail recovery. Our method involves extracting semantic data, including textual descriptors and spatial positioning, and encoding this information. We also present a Cross-Modal Feature Fusion module for generating semantic feature maps via multimodal training, and a Prompt Traction Head to augment the backbone network with semantic insights, thereby refining image details. Our empirical results across various datasets validate the superiority of our approach.
\end{abstract}

\vspace{0.5em}

\section{Introduction}

Image super-resolution, a classic computer vision task, upscales low-resolution images to high-resolution ones without information loss \cite{cite-01}. It finds applications in image compression transmission, detail enhancement, and defect restoration \cite{newcite-01}. Recently, Convolutional Neural Network (CNN) and transformer-based methods have been proposed \cite{cite-06, cite-07, cite-08, cite-09, cite-10, cite-11, cite-13, cite-14, cite-15}, exploiting their feature extraction capabilities for image detail restoration. However, these methods overlook the rich semantic information within images, which, if exploited, could significantly improve output quality.

This paper proposes a novel approach that leverages semantic image information using models like YOLO \cite{wcite-01} and BLIP \cite{newcite-16}. The extracted information is incorporated into the super-resolution model's training process as a prompt. Additionally, an attention mechanism is employed to apply the encoded information to specific image areas, reflecting the regional properties of image semantics.

\begin{figure}[htbp]
 \begin{minipage}[t]{0.5\linewidth}
		\centering
		\includegraphics[width=0.9\linewidth]{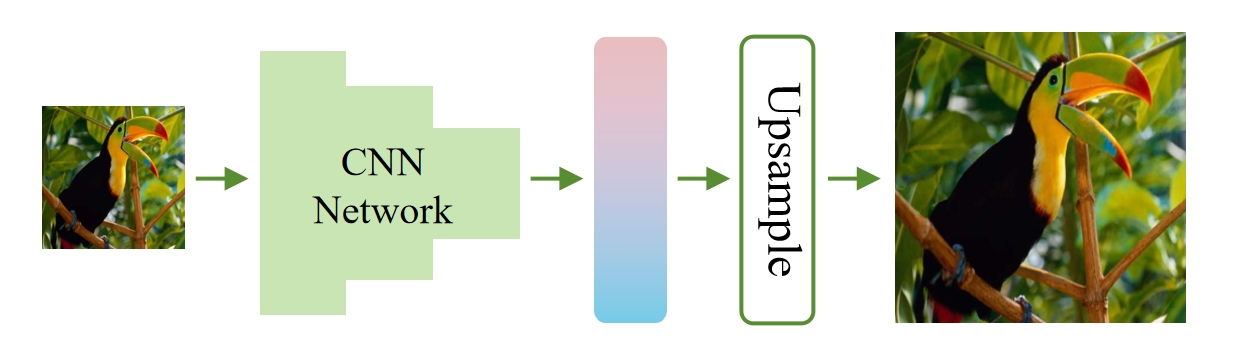}
		\subcaption*{\linespread{0.5}\selectfont (a) Classic CNN-based image super-resolution models.}
		\label{fig:intro:b}
 \end{minipage}
\begin{minipage}[t]{0.5\linewidth}
		\centering
		\includegraphics[width=0.9\linewidth]{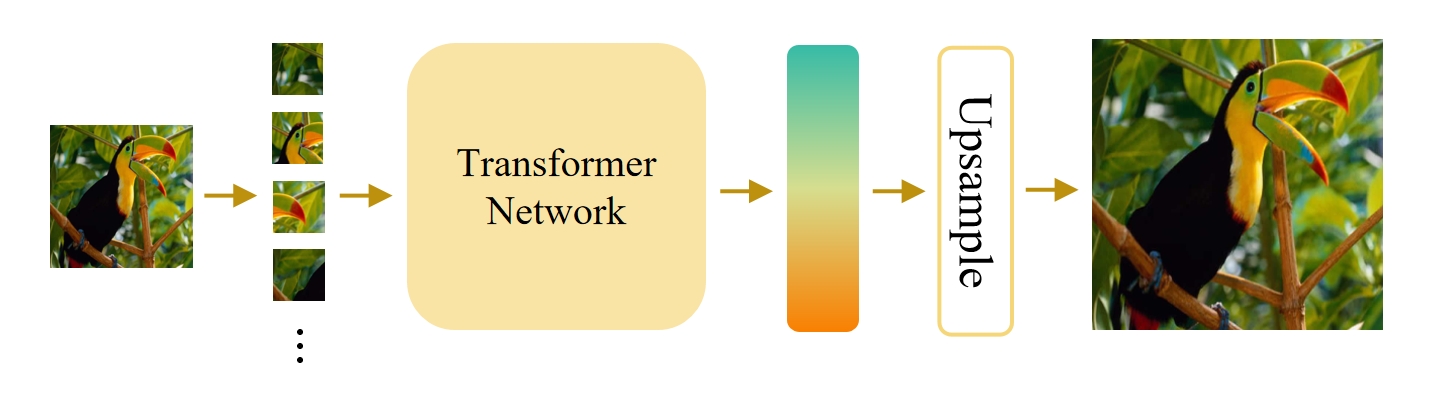}
		\subcaption*{\linespread{0.5}\selectfont (b) Transformer-based image super-resolution model.} 
		\label{fig:intro:c}
\end{minipage}
\begin{minipage}[t]{0.5\linewidth}
		\centering
		\includegraphics[width=0.9\linewidth]{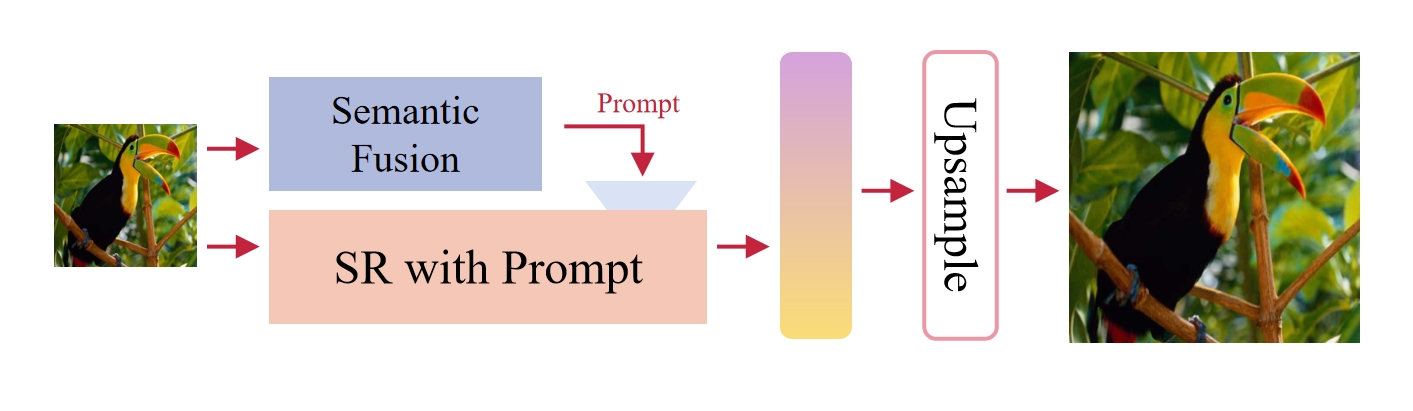}
		\subcaption*{\linespread{0.5}\selectfont (c) Proposed image super-resolution model.}
		\label{fig:intro:d}
	\end{minipage}
\begin{minipage}[t]{0.5\linewidth}
		\centering
		\includegraphics[width=0.9\linewidth]{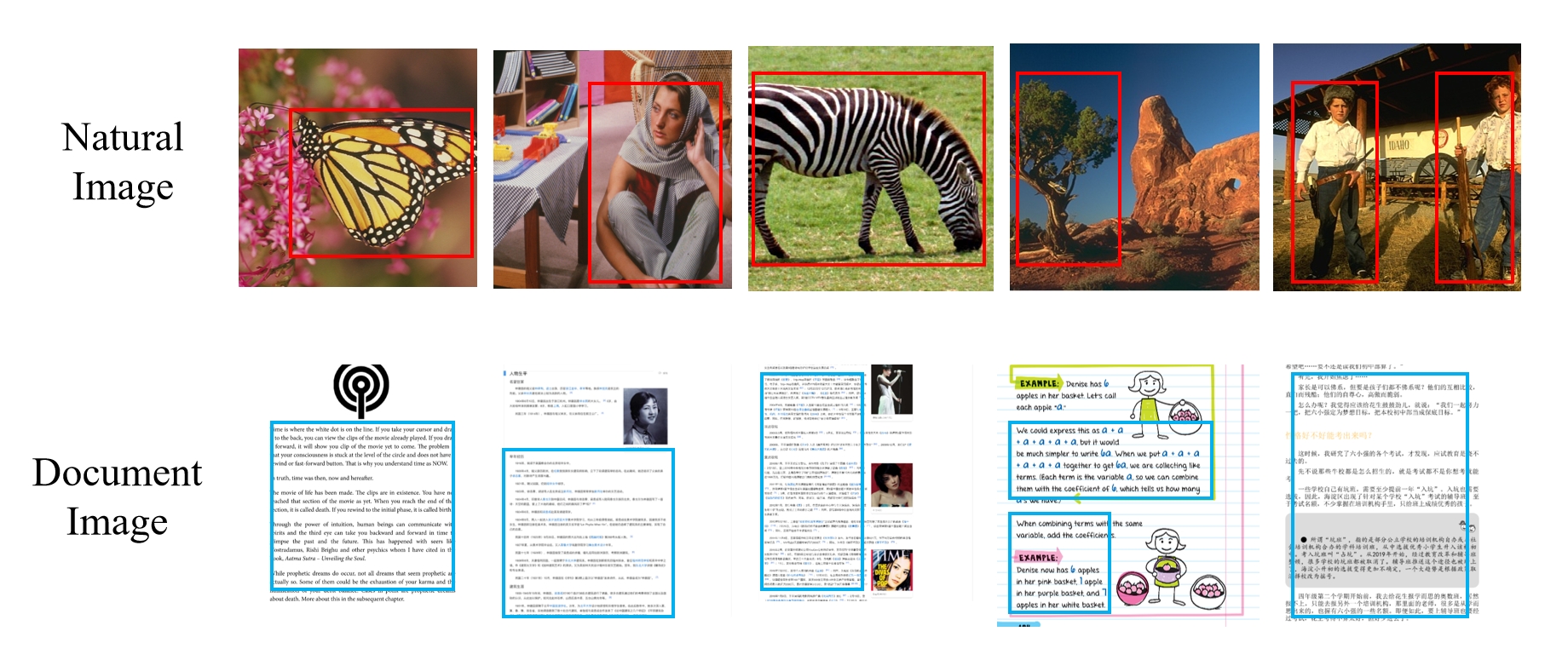}
		\subcaption*{\linespread{0.5}\selectfont (d) Key content areas for different image types.} 
		\label{fig:intro:a}
	\end{minipage}
\caption{Comparison of different image super-resolution model architectures.}	
\label{fig:intro}
\end{figure}

Figure \ref{fig:intro} illustrates the difference between our approach and traditional super-resolution methods. Traditional CNN-based methods (Figure \ref{fig:intro}(a)) extract features at different scales, which are then fused and upsampled to restore image details \cite{cite-09}. Transformer-based methods (Figure \ref{fig:intro}(b)) divide the image into patches, transform them into fixed-length vectors, and process them through self-attention mechanisms \cite{wcite-03, wcite-04}. Our approach (Figure \ref{fig:intro}(c)) incorporates semantic extraction and cross-modal feature fusion modules, extracting textual semantic information from the image and encoding it into feature vectors that serve as prompts for the super-resolution model.

Semantic information varies across image categories due to content differences (Figure \ref{fig:intro}(d)). For natural images, the main content may consist of objects, people, animals, etc., while document images primarily consist of text with substantial blank areas between paragraphs \cite{wcite-05}. Therefore, we employ an Image Caption model \cite{wcite-06} and an Object Detection model \cite{wcite-07} for natural images, and an OCR model \cite{wcite-08} for document images.

We conducted extensive experiments to validate our approach, comparing it with several state-of-the-art super-resolution algorithms. The results demonstrate that our method effectively utilizes image semantic information, improving various performance metrics and visual effects of the model.

In summary, this paper contributes to image super-resolution by proposing a novel method that utilizes image semantic information for model training, improving model performance. We design different schemes for extracting semantic information for different types of images and establish associations between the extracted semantic information and the corresponding image area. Extensive experiments validate the proposed method, demonstrating its effectiveness in utilizing semantic information and improving various performance metrics and visual effects.

\section{Related Work}

\subsection{Single Image Super-Resolution}

Deep learning-based methods, such as SRCNN \cite{cite-09}, which first applied convolutional neural networks (CNN) to image super-resolution tasks, are commonly used in this field. Subsequent methods \cite{cite-06,cite-07,cite-08,cite-09,cite-10,cite-11,cite-13,cite-14,cite-15} improved image reconstruction quality by optimizing convolutional modules and incorporating adversarial learning \cite{cite-27,cite-28}. Recent Transformer-based networks \cite{cite-13,cite-14,cite-15} have shown excellent results due to their powerful self-attention mechanisms. 

Image super-resolution methods have been applied to specific image types, such as medical \cite{newcite-04}, hyperspectral \cite{newcite-06}, and scene images \cite{newcite-07}. These methods often struggle with generalization due to their optimization for specific image types. Our method overcomes this limitation by using different semantic extraction modules for different image types, leading to stronger generalization performance.

\subsection{Text Encoding}

Text encoding in Natural Language Processing (NLP) transforms textual data into a numerical format for machine learning algorithms \cite{cite-45}. Key techniques include Bag-of-Words (BoW) Encoding \cite{wcite-09}, Word Embeddings (e.g., Word2Vec \cite{wcite-10}, GloVe \cite{wcite-11}, FastText \cite{wcite-12}), and Transformers and Pretrained Language Models (e.g., BERT \cite{cite-51} and GPT \cite{cite-50}). We utilize BERT \cite{cite-51} for its superior performance in text encoding tasks.

\subsection{Feature Fusion}

Feature fusion integrates features from different modalities to improve cross-modal task performance. Techniques include early fusion \cite{wcite-15}, which combines features at a low level, late fusion \cite{wcite-16}, which combines features at a higher semantic level, and hybrid fusion \cite{wcite-17}, which integrates features at multiple stages.

\subsection{Image Semantic Extraction}

Our method utilizes a combination of Object Detection, Image Captioning, and Optical Character Recognition (OCR) to extract semantic information from images. For natural images, we use the YOLOv5 model \cite{yolov5} for object detection and BLIP \cite{newcite-16} to extract deeper semantic information. For document images, PaddleOCR \cite{wcite-18} extracts semantic information with high accuracy.

\section{Methodology}

\subsection{Defining the Problem}

The primary objective of Image Super-Resolution (SR) is to augment the resolution of a low-resolution (LR) image, thereby deriving a high-resolution (HR) image replete with intricate details. From a mathematical perspective, the super-resolution problem can be defined as follows:

Assume an input LR image $\mathbf{X} \in \mathbb{R}^{m \times n}$, where $m$ and $n$ denote the spatial dimensions. The aim is to create an HR image $\hat{\mathbf{Y}} \in \mathbb{R}^{p \times q}$, where $p$ and $q$ are the upsampled dimensions.

This relationship between the LR image and HR image can be formally expressed as: \begin{equation*} \mathbf{X} = \mathcal{D}(\mathbf{Y}) + \mathbf{N}, \end{equation*} where $\mathbf{Y} \in \mathbb{R}^{p \times q}$ signifies the original HR image, $\mathcal{D}(\mathbf{Y})$ symbolizes the degradation operator which models the downsampling process, and $\mathbf{N}$ represents the additive noise. The objective of the image super-resolution algorithm is to ensure that $\hat{\mathbf{Y}}$ closely approximates $\mathbf{Y}$, thus achieving a high degree of similarity.

The quality of the reconstructed HR image can be assessed using various image quality metrics such as the peak signal-to-noise ratio (PSNR) or the structural similarity index (SSIM) \cite{newcite-21}.

However, due to the inherent ill-posed nature of the problem, a unique solution is elusive, thereby making the super-resolution problem a challenging one. To overcome this issue, a variety of approaches have been suggested, encompassing both traditional optimization-based methods and methods based on deep learning.

In this study, we put forth a super-resolution method that integrates semantic information extracted from images as cues during the model training phase. To ascertain its efficacy, we juxtapose our approach with cutting-edge methods on benchmark datasets. The reconstructed HR images are assessed using quantitative metrics. Experimental results reveal that by capitalizing on semantic prompts, our method surpasses the performance of existing algorithms.

\subsection{Overview of the Method}

Our proposed method for image super-resolution strives to utilize both pixel features and semantic features to achieve results that are not only more accurate but also visually appealing. The fundamental concept is to incorporate the extracted semantic information as cues during the model training phase, thereby enabling the model to be directed by the context and content of the images.

\begin{figure}[ht]
\centering
\includegraphics[width=\linewidth]{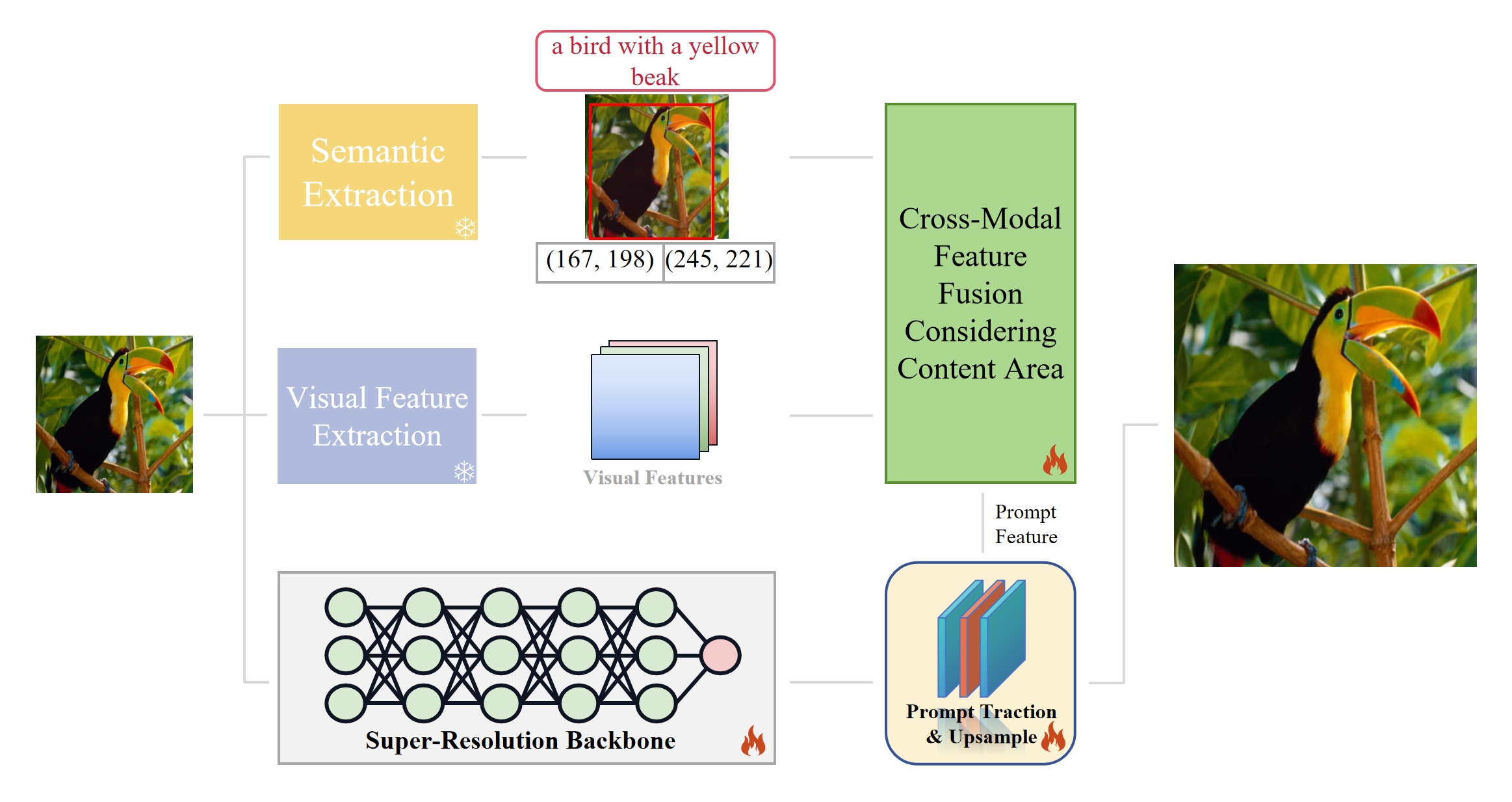}
\caption{The comprehensive framework of the proposed method. The snowflakes in the figure denote that the module is in a frozen state, while the flames indicate that the module is undergoing training.}
\label{fig:framework}
\end{figure}

The overall workflow of our method is illustrated in Figure \ref{fig:framework}. It comprises a series of components: 

\begin{itemize}
\item Super-Resolution Backbone
\item Semantic Extraction \& Visual Extraction
\item Cross-Modal Feature Fusion
\item Prompt Traction Head
\end{itemize}

The \textbf{Super-Resolution Backbone} is crucial for preserving image details and textures during the super-resolution process. It utilizes advanced algorithms and deep learning to map low-resolution to high-resolution spaces. We adopt the ESRGAN generator from \cite{cite-08} as our backbone, a UNET-like architecture that is highly compatible with complex image data. ESRGAN is extensively employed in various computer vision tasks such as image segmentation, making it a dependable choice for super-resolution. Its forte lies in enhancing image details and high-frequency information, which are vital for improving image quality in super-resolution tasks.

The \textbf{Semantic Extraction \& Visual Extraction} component is primarily responsible for extracting the semantic information and visual features of an image. In our method, we use a pre-trained VGG16 \cite{wcite-19} to extract visual features of images for subsequent cross-modal feature fusion. For image semantic information extraction, we use BLIP \cite{newcite-16} and YOLOv5 \cite{yolov5} for semantic extraction of natural images, and PaddleOCR \cite{wcite-18} for semantic extraction of document images. The extracted semantic information consists of two parts: text description and region location coordinates. The combination of text description and location information forms a natural language representation of key regions in the image, which can be encoded and utilized for model training through language modeling.

The \textbf{Cross-Modal Feature Fusion} component is used to align and optimize visual and textual modal features, enabling the model to learn more profoundly from multimodal features. Moreover, our method is designed to effectively utilize semantic region position information, which includes encoding location in the form of coordinates and involving location features in training. By doing so, the model is able to consider different regions in the image separately, thus improving the performance. In terms of concrete implementation, we leverage the self-attention mechanism proposed by Transformer to design a multimodal attention module that considers position information for image super-resolution tasks.

The \textbf{Prompt Traction Head} component guides and optimizes the output of a super-resolution network by inputting the prompt feature obtained from multimodal feature fusion. Specifically, we use the prompt feature as a kind of exciters to increase or suppress the values of the image feature maps, allowing the details of the feature maps to be better displayed. The processed feature maps are then processed by the upsampling module to obtain the enlarged image.

\subsection{Semantic Processing}

This section delves into the generation and processing of semantic information in our proposed approach. We utilize a fully trained, domain-specific model to automatically extract semantic information from images. This extracted information is then transformed into textual data for feature extraction, deploying a well-established language model, as detailed in \cite{cite-50, cite-51}. Recognizing the importance of localized semantic content within an image, we also consider the spatial distribution of key elements.

It's crucial to acknowledge that semantic information varies across different image types. In natural images, semantic information is typically the types of objects in the image and their spatial relationships. However, for document images, it refers to the textual content within the image. Hence, we have specifically defined the input and output interfaces for the semantic extraction module, thereby enhancing the flexibility of our approach. The module receives an image as input and outputs a textual representation of the image's semantics and the coordinates of significant regions. We have developed specific semantic extraction modules for both natural and document images, which have been utilized in our subsequent experiments.

\begin{figure}[htb]
	\begin{minipage}[t]{1.0\linewidth}
		\centering
		\includegraphics[width=0.9\linewidth]{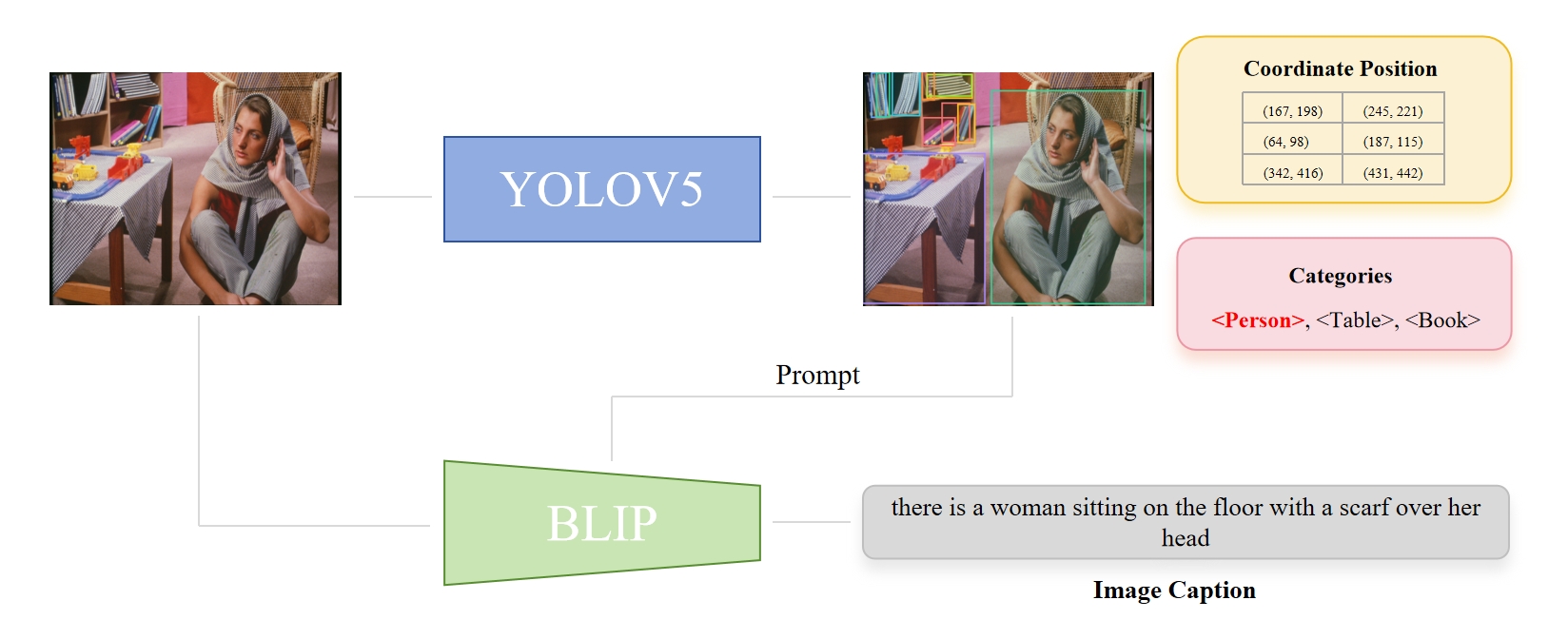}
		\subcaption*{\linespread{0.5}\selectfont (a) For natural images, we employ YOLOv5 to identify the category and position of objects in the image, and utilize BLIP to generate a textual description of the image based on the detected objects' names.} 
		\label{fig:extraction:a}
	\end{minipage}
    \vspace{15pt}
	\begin{minipage}[t]{1.0\linewidth}
		\centering
		\includegraphics[width=0.9\linewidth]{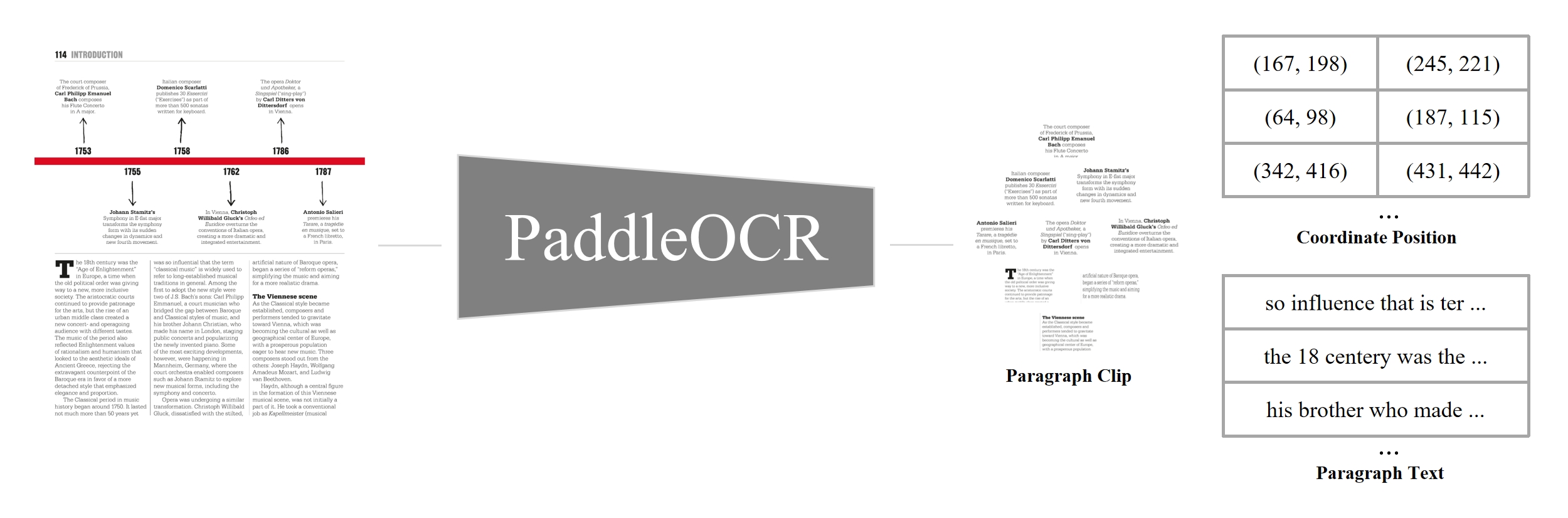}
		\subcaption*{\linespread{0.5}\selectfont (b) For document images, we directly apply PaddleOCR to extract the positions and text content of paragraphs in the image.}
		\label{fig:extraction:b}
	\end{minipage}
\caption{Semantic Information Processing for Different Types of Images.}	
\label{fig:extraction}
\end{figure}

As depicted in Figure \ref{fig:extraction}(a), for natural images, our process begins with YOLOv5 \cite{yolov5} to detect objects, determining their category and spatial information. The image is then processed by BLIP \cite{newcite-16} to create a textual description, using the detected objects' categories as prompts to guide text generation. Our object detection model excels at recognizing diverse object categories. Thus, when selecting prompt words, we prioritize categories based on the area they occupy in the image, opting for those with the greatest coverage. This strategy enhances the creative capability of the image captioning model.

For document images, as shown in Figure \ref{fig:extraction}(b), we use the PaddleOCR model \cite{wcite-18} to identify text content and its corresponding spatial coordinates. As document images primarily contain textual elements arranged in coherent paragraphs, we argue that these paragraphs inherently carry the semantic essence of such images. Based on this premise, we directly apply an OCR model to extract both the textual content and the precise positioning of paragraph regions. This methodology ensures consistency in our semantic extraction across various image types.

\subsection{Cross-Modal Feature Fusion}

The Cross-Modal Feature Fusion module is designed to integrate features from different sources, namely visual and semantic domains. The inputs to this module are detailed as follows:

\begin{itemize}
\item \textbf{Visual Features:} These are the features extracted from the image using a convolutional neural network (CNN). The visual features are high-dimensional representations that capture the visual content of the image, including textures, edges, and color information. The dimensionality and the level of abstraction of these features depend on the depth of the CNN and the layer from which they are extracted.

\item \textbf{Positional Masks:} Positional masks are binary or continuous maps that provide spatial information about regions of interest within the image. These masks are used to highlight or suppress specific areas during the feature fusion process, allowing the model to focus on salient parts of the image that are more relevant to the task at hand.

\item \textbf{Semantic Features:} Semantic features are derived from the textual information associated with the image. These features are obtained using a language model such as BERT, which encodes the textual semantics into a high-dimensional vector space. The semantic features capture the contextual information and the relationships between different elements described in the text. Positional information is also encoded and added to the semantic features to ensure that the spatial relationships within the image are preserved during the fusion process.

\end{itemize}

Once these inputs are prepared, the Cross-Modal Feature Fusion module proceeds to perform a series of operations to merge the visual and semantic features effectively. 

\subsubsection{Semantic Feature Preparation}

In image semantic extraction, the primary focus often revolves around analyzing specific localized regions. However, for model training efficiency, it is vital to integrate positional information into the semantic features. To this end, we have developed an innovative position encoding module that effectively incorporates positional context into semantic features. Additionally, in Equation \ref{eq:loss}, we have utilized cosine similarity as a crucial part of the loss function. This strategic decision enables the model to skillfully uncover the relationship between visual and semantic cues while concurrently maintaining the coherence of positional insights across diverse modalities.

\begin{figure}[htbp]
	\begin{minipage}[t]{1.0\linewidth}
		\centering
		\includegraphics[width=0.9\linewidth]{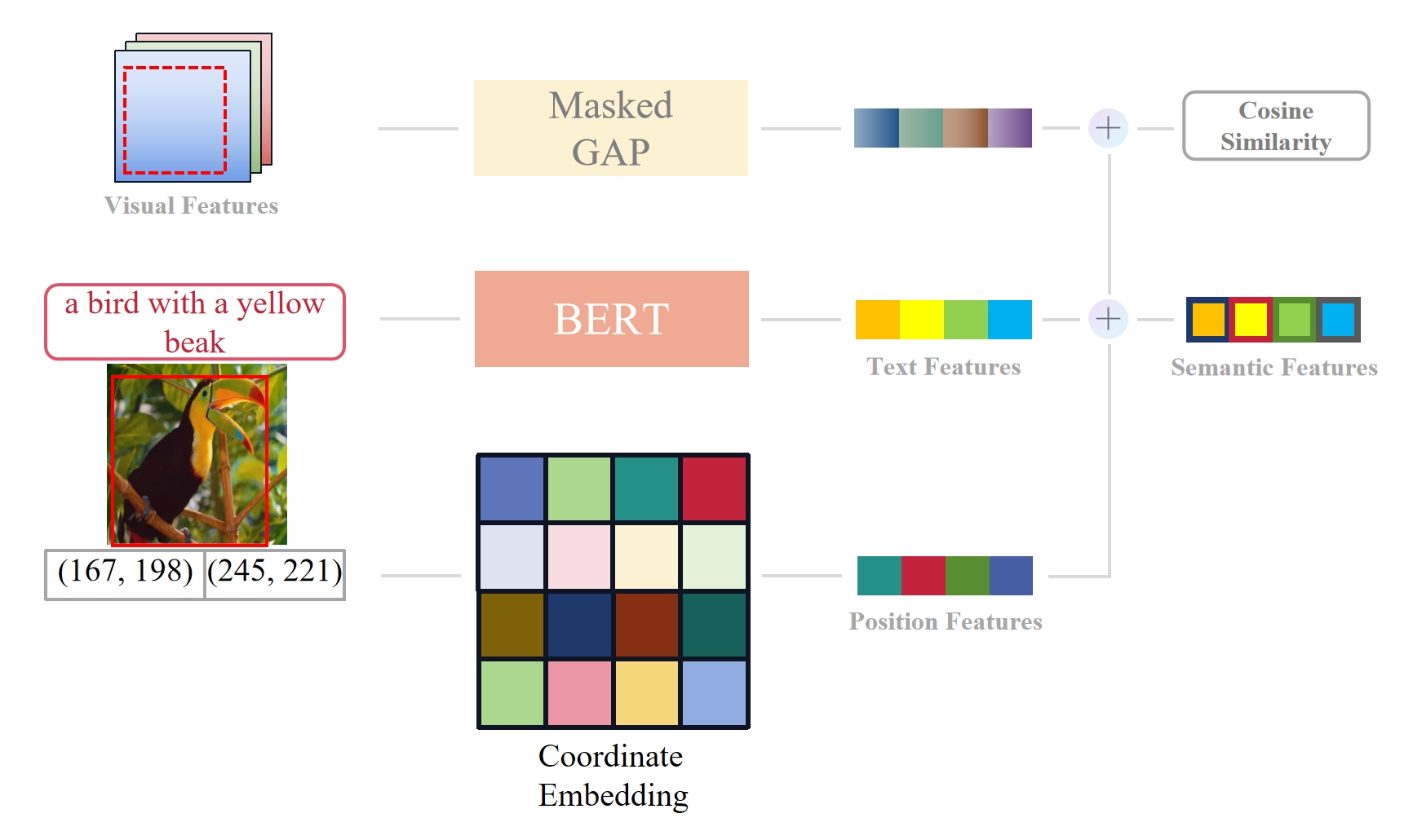}
		\subcaption*{\linespread{0.5}\selectfont (a) Semantic feature generation. GAP operations are performed on visual features based on the coordinates of the content area. Positional information is encoded and added to multimodal features. To maintain semantic consistency between visual and textual features, a cosine loss function is used for constraints.} 
		\label{fig:fuse:a}
	\end{minipage}
    \vspace{15pt}
	\begin{minipage}[t]{1.0\linewidth}
		\centering
		\includegraphics[width=0.9\linewidth]{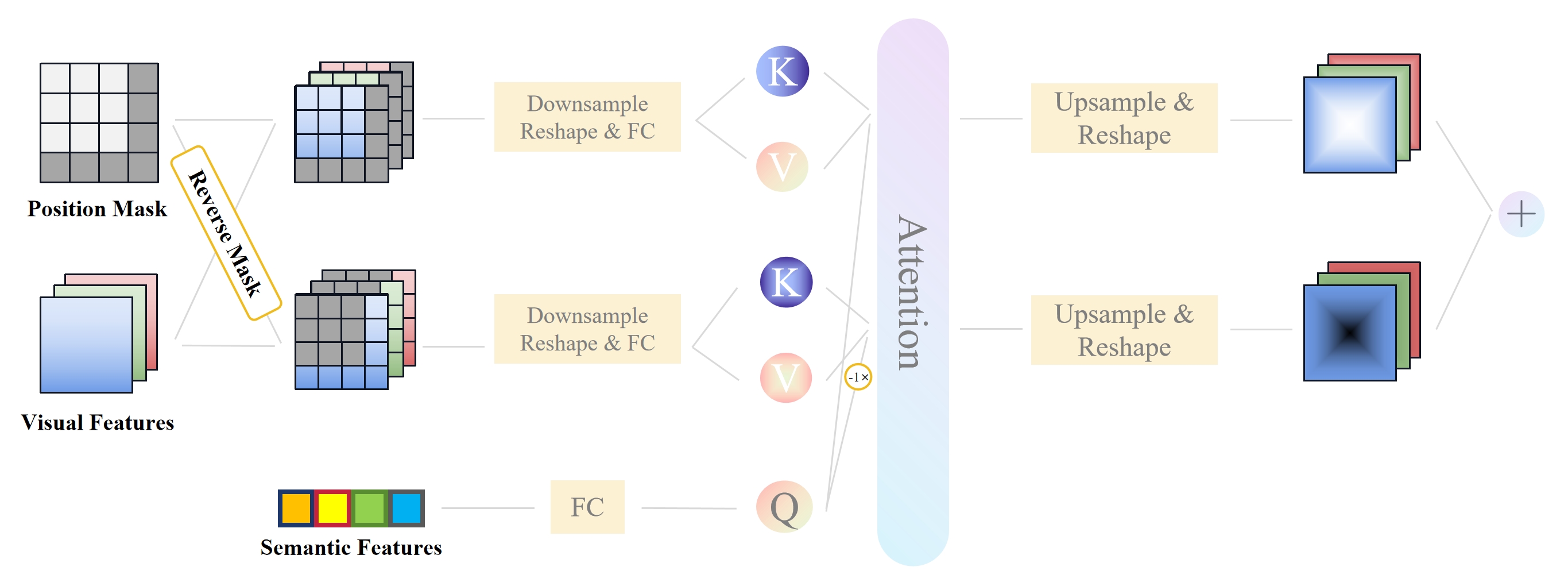}
		\subcaption*{\linespread{0.5}\selectfont (b) Cross-modal feature fusion. By integrating the position information of the content area, attention operations are performed on features of different modalities for feature fusion. Content and non-content regions are added after parallel operations.}
		\label{fig:fuse:b}
	\end{minipage}

\caption{Structure of Position Mask Attention.}	
\label{fig:fuse}
\end{figure}

The comprehensive workflow is detailed in Figure \ref{fig:fuse}(a). The textual semantics extracted from the image are encoded using BERT \cite{cite-51}, a state-of-the-art language model. Simultaneously, we use an efficient matrix embedding layer to encode positional data, adeptly converting coordinates into high-dimensional vectors. Unlike traditional positional encoding methods, such as those used in Vision Transformer (VIT) \cite{cite-64}, our approach represents information in the form of two-dimensional coordinates. When processing visual features, the first step involves generating a mask derived from positional data. Then, the positional features are smoothly fused with their corresponding visual counterparts based on the underlying mask. The generation of local visual features within specific regions is achieved using the Masked Global Average Pooling method \cite{cite-65}.

Upon completion of semantic feature generation as depicted in Figure \ref{fig:fuse}(a), the output of the image semantic extraction module is a multi-dimensional vector. This vector encapsulates the semantic information derived from the image, which has been processed through the BERT language model. The dimensions of the output vector are determined by the architecture of BERT and the specific implementation details, typically resulting in a fixed-size embedding that represents the semantic content. For instance, if BERT-Base is used, the output dimension would be 768, corresponding to the hidden size of the BERT-Base model. This semantic feature vector is then ready to be fused with the visual features, which have been similarly encoded into a high-dimensional space reflective of the positional information and visual content of the image.

It is important to note that the dimensionality of the semantic features must be compatible with the visual features for effective fusion. If there is a discrepancy in dimensions, a transformation operation, such as a fully connected (FC) layer, may be employed to align the semantic features with the visual features. This ensures that the subsequent position mask attention module can effectively perform its task, leveraging both semantic and visual cues to enhance the model's understanding of the image content.

\subsubsection{Position Mask Attention}

Traditional multimodal feature fusion primarily relies on Transformer-based processing, as demonstrated by previous works like Transformer \cite{cite-12}. Given that feature extractors for various modalities all adopt the Transformer architecture, a straightforward approach involves summing these features and further processing them to facilitate multimodal training. However, when dealing with convolutional neural networks, the fusion of distinct modal features becomes a more complex challenge \cite{newcite-18}. Merely adjusting dimensions and performing summations often fail to achieve the desired synergistic effect. Therefore, we incorporate an attention mechanism \cite{newcite-23} to unite diverse modal features while simultaneously distinguishing between critical and non-critical regions within the image.

As shown in Figure \ref{fig:fuse}(b), this module takes visual features, position masks, and semantic features as input and proceeds to outline discrete regions within the visual features with the help of masks.

\begin{align}
    F_{mask} &= F_{visual} \times X_{mask} \\ 
    F_{mask}^{-1} &= F_{visual} \times X_{mask}^{-1}
\end{align}

Given the size disparity within the visual features, we first initiate a $2\times$ downsampling of both \texttt{$F_{mask}$} and \texttt{$F_{mask}^{-1}$}, followed by a reflective transformation applied to all features. In the subsequent equation, the variables $\theta$, $\delta$, and $\lambda$ are employed as learnable parameters.

\begin{align}
    K_{\alpha }, V_{\alpha } &= FC(\theta ,F_{mask}) \\
    K_{\beta }, V_{\beta } &= FC(\delta ,F_{mask}^{-1}) \\ 
    Q &= FC(\lambda, F_{semantic})
\end{align}

Next, we use the semantic features as queries and perform attention operations on the key and non-key regions, distinctly facilitating the fusion of different modal features.

\begin{align}
    F &= \sum_{i}^{\alpha ,\beta }Upsample(\omega_{i}(Q^{*} \otimes K_{i}^{T}) \times V_{i} ) \\
    Q^{*} &= \left\{\begin{matrix} Q \quad &i=\alpha \\-1\times Q \quad &i=\beta \end{matrix}\right.
\end{align}

To conclude the process, the fused features undergo dimensional transformation and upscaling, resulting in the summation of features from different regions. It is worth emphasizing that we designate $Q$ as the semantic features related to critical regions. In our effort to make the features in non-critical regions distinctly separable from those in critical regions, we use a negated version of $Q$ (-$Q$) for the attention operation within non-critical regions.

\subsection{Prompt Traction Head}

The Prompt Traction Head is tasked with directing and optimizing the super-resolution network's output by utilizing prompt features derived from multimodal feature fusion. This section delves into the specific operations within the Prompt Traction Head, including the enhancement of image feature details using prompts and the subsequent process of image upscaling.

In particular, prompt features serve as exciters that modulate the values within image feature maps, thereby bolstering the depiction of feature details. These feature maps are then subjected to additional processing via an upsampling module, resulting in the final upscaled image.

The operations of the Prompt Traction Head can be expressed through the following equations:

\begin{align}
    F_{\text{excited}} &= Conv(F_{\text{prompts}}) + F_{\text{sr}} \cdot \delta(F_{\text{prompts}}) \\ 
    F_{\text{upscaled}} &= \text{Upsample}(F_{\text{excited}})
\end{align}

Here, $\delta$ denotes the $tanh$ function and $Conv$ symbolizes the point convolution operation. $F_{\text{sr}}$ symbolizes the feature map produced by the super-resolution backbone, while $F_{\text{prompts}}$ signifies the prompt feature post cross-modal fusion. The incorporation of prompts within this module enables a more nuanced manipulation of the feature maps, ensuring that the image's fine-grained details are effectively preserved and amplified during the super-resolution process. This method yields visually appealing and precise super-resolved images.

\begin{figure}[ht]
	\begin{minipage}[t]{1.0\linewidth}
		\centering
		\includegraphics[width=0.6\linewidth]{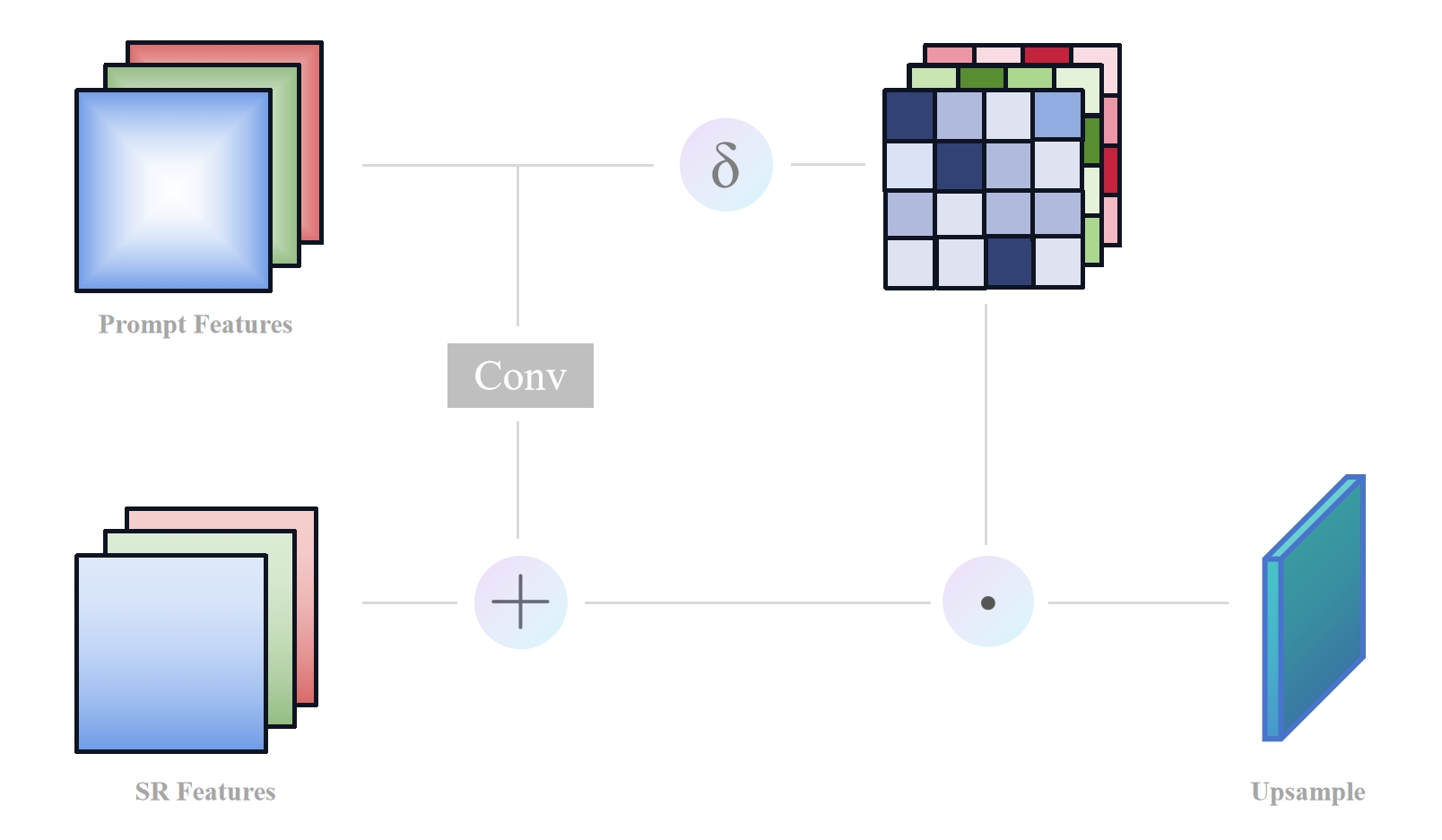}
		\label{fig:head:a}
	\end{minipage}
\caption{The structural diagram of the Prompt Traction Head, where $\delta$ denotes the $tanh$ function and $Conv$ symbolizes the point convolution operation.}	
\label{fig:head}
\end{figure}

As depicted in Figure \ref{fig:head}, our design is characterized by its simplicity and efficacy. Initially, a point convolution is applied to map $F_{\text{prompts}}$ once, followed by an element-wise addition to merge it with $F_{\text{sr}}$. This operation aims to integrate the image's semantic information with the original features, thereby augmenting the expressive capacity of these features. Simultaneously, $F_{\text{prompts}}$ undergoes a $tanh$ transformation to generate parameter values for modulating the excitation intensity of $F_{\text{sr}}$. This modulation process, which involves element-wise multiplication, enables the amplification or suppression of the integrated semantic information's features. Ultimately, a straightforward deconvolution operation is employed to carry out the upsampling, culminating in the creation of an upscaled image.

\subsection{Metric and Loss Function}

In our methodology, we deploy a carefully curated set of metrics and loss functions, which serve as the guiding light for the super-resolution process. We incorporate traditional image quality metrics, such as PSNR and SSIM \cite{newcite-21}, to provide a quantitative assessment of the fidelity of the super-resolved images.

\begin{align}
\label{eq:loss}
\mathcal{L}(x, y, F_{vis}, F_{sem}, F_{sr} ) 
& = L1(x, y) 
+ \alpha \cdot \mathcal{S}(F_{sem} , F_{vis}) 
+ \beta \cdot \mathcal{V}(F_{sr}, F_{vis}) 
\\
L1(x, y) & = \left \|  x - y  \right \|_{1}
\\
\mathcal{S}(F_{sem} , F_{vis})  &= cosine(F_{sem} - GAP(F_{vis})) 
\\ 
\mathcal{V}(F_{sr}, F_{vis})   &= \sqrt{(  F_{sr} - MaxPool_{\times scale}(F_{vis})  )^{2}}
\end{align}

We have exercised considerable care in the design of our loss function. As depicted in the formula above, our loss function integrates three critical components: pixel loss, semantic similarity loss, and visual feature similarity loss. In this equation, $\alpha$ and $\beta$ denote weight coefficients, $F_{vis}$ corresponds to the visual features of the generated image, $F_{sem}$ signifies the semantic features of the generated image, and $F_{sr}$ represents the features derived from super-resolution backbone downsampling. The pixel loss employs an L1 loss \cite{newcite-20} function to optimize the distance between the generated image and the real image. Semantic similarity loss is harnessed to ensure the consistency between image features and semantic features. Visual feature similarity loss is utilized to maintain the expressive resemblance between visual feature extractors and the super-resolution backbone at the feature level. It is noteworthy that, due to the inconsistency in the size of the feature maps generated by the two networks, we employ a MaxPool operation to standardize the size, thereby facilitating the execution of calculations.

\subsection{Discussion on Computational Costs}

Our semantic processing method is effective for image understanding but computationally demanding. The YOLOv5 model offers a good speed-accuracy tradeoff for object detection but is resource-intensive, especially with high-resolution images on limited-GPU systems.

The BLIP model for textual descriptions also requires significant GPU resources due to its advanced vision-language architecture. It can be taxing when producing detailed captions.

Extracting text and coordinates from document images with PaddleOCR can become costly for high-resolution, text-dense images, leading to increased processing times and memory demands.

Additionally, BERT's deep bidirectional architecture for text feature extraction further adds to the computational load. Its thorough language analysis is resource-heavy, potentially slowing down real-time applications or large dataset processing.

\section{Experimental Procedures}

\subsection{Configuration of the Experiment}

\textbf{Datasets.} Our experiment utilizes two datasets: the DF2K dataset (DIV2K \cite{cite-67} combined with Flickr2K \cite{cite-68}), and the Baidu Document Image dataset (BDDI) \cite{cite-66}. These datasets serve as the training sets, enabling us to examine the efficacy of our method on both natural and document images. For our super-resolution network, we employ the generator component of ESRGAN, incorporating a UNET-inspired network architecture that has proven to be effective in handling complex image data. To provide a comparative analysis, we adjusted the number of network layers and channels in ESRGAN \cite{cite-08}, PAN \cite{cite-07}, REALSR \cite{cite-06}, SWINIR \cite{cite-14}, and HAT \cite{cite-15} to maintain a similar parameter count across all models. It is important to note that none of these models were pre-trained using any other datasets.

\textbf{Implementation Details.} We configured the language model to have a maximum text input length of 128 and a text feature dimensionality of 768. The dimensionality remains the same after position information has been encoded. In the cross-modal attention mechanism, we downsample the feature map and map it through the MLP layer, specifying a feature dimension of 256 after mapping. We discovered through experimentation that optimal model performance can be achieved by adjusting the mask magnitude in the content region masking operation, which we set to 0.3. To investigate the model's performance at different image scales, we conducted image super-resolution experiments at $2\times$, $3\times$, and $4\times$. The performance of the model was evaluated using PSNR and SSIM metrics \cite{newcite-21}. All computations in the experiments were performed using two NVIDIA V100 GPUs.

\subsection{Enhanced Performance Comparison with State-of-the-Art Methods}

\subsubsection{Evaluation on Natural Image Datasets}

\begin{table}[htb]
\centering
\caption{Comparative Results on Natural Image Datasets. Our method surpasses other techniques in numerous benchmarks, particularly leveraging ESRGAN as the backbone while achieving significant improvements over it.}
\label{tab:nature}
\resizebox{\textwidth}{!}{%
\begin{tabular}{lcccccc}
\hline
Method & Scale     & Set5                    & Set14                   & BSD100                  & Urban100                & Manga109                \\ \hline
ESRGAN & $\times$2 & 37.96 / 0.9593          & 33.87 / 0.9135          & 31.62 / 0.8902          & 32.27 / 0.9307          & 39.56 / 0.9648          \\
PAN    & $\times$2 & 38.34 / 0.9625          & 34.36 / 0.9221          & 32.00 / 0.9001          & 33.42 / 0.9402          & 39.84 / 0.9726          \\
RealSR & $\times$2 & 38.19 / 0.9622          & 34.38 / 0.9223          & 31.98 / 0.8967          & 33.89 / 0.9395          & 39.79 / 0.9721          \\
SwinIR & $\times$2 & 38.37 /  0.9617         & 34.42 / 0.9228          & 32.02 / 0.9000          & 33.26 / 0.9387          & 39.76 / 0.9710          \\
HAT    & $\times$2 & 38.59 / 0.9629          & 34.47 / 0.9223          & \textbf{32.12 / 0.9009} & 33.44 / 0.9418          & 40.02 / 0.9729          \\
Ours   & $\times$2 & \textbf{38.62 / 0.9636} & \textbf{34.55 / 0.9249} & 32.10 / 0.9024          & \textbf{33.60 / 0.9425} & \textbf{40.14 / 0.9730} \\ \hline
ESRGAN & $\times$3 & 35.24 / 0.9347          & 30.76 / 0.8557          & 29.44 / 0.8607          & 29.84 / 0.8803          & 36.59 / 0.9471          \\
PAN    & $\times$3 & 35.41 / 0.9351          & 30.80 / 0.8571          & 29.44 / 0.8610          & 29.95 / 0.8816          & 36.67 / 0.9488          \\
RealSR & $\times$3 & 35.39 / 0.9360          & 30.79 / 0.8560          & 29.49 / 0.8606          & 29.89 / 0.8805          & 36.70 / 0.9491          \\
SwinIR & $\times$3 & 36.11 /  0.9445         & 31.52 / 0.8671          & 29.66 / 0.8611          & 30.49 / 0.8837          & 36.90 / 0.9531          \\
HAT    & $\times$3 & \textbf{36.20 / 0.9444} & 31.57 / 0.8669          & \textbf{29.71 / 0.8613} & 30.55 / 0.8844          & 36.98 / 0.9529          \\
Ours   & $\times$3 & 36.15 / 0.9441          & \textbf{31.63 / 0.8674} & 29.71 / 0.8615          & \textbf{30.57 / 0.8849} & \textbf{37.05 / 0.9537} \\ \hline
ESRGAN & $\times$4 & 32.73 / 0.8915          & 27.72 / 0.7794          & 27.64 / 0.7406          & 26.48 / 0.8158          & 31.62 / 0.9146          \\
PAN    & $\times$4 & 32.58 / 0.8881          & 27.73 / 0.7770          & 27.66 / 0.7498          & 26.51 / 0.8172          & 31.77 / 0.9173          \\
RealSR & $\times$4 & 32.59 / 0.8879          & 27.86 / 0.7782          & 27.66 / 0.7462          & 26.49 / 0.8161          & 31.79 / 0.9169          \\
SwinIR & $\times$4 & 32.66 / 0.8889          & 28.59 / 0.7850          & 27.69 / 0.7422          & 27.50 / 0.8260          & 31.85 / 0.9184          \\
HAT    & $\times$4 & 32.77 / 0.8912          & 28.80 / 0.7805          & 27.71 / 0.7475          & \textbf{27.63 / 0.8227} & 32.07 / 0.9222          \\
Ours   & $\times$4 & \textbf{32.84 / 0.8947} & \textbf{28.87 / 0.7830} & \textbf{27.76 / 0.7481} & 27.43 / 0.8269          & \textbf{32.13 / 0.9234} \\ \hline
\end{tabular}%
}
\end{table}

Our model was meticulously trained on the expansive DF2K dataset and subsequently subjected to comprehensive evaluations on several benchmark datasets, including Set5 \cite{cite-71}, Set14 \cite{cite-72}, BSD100 \cite{cite-73}, Urban100 \cite{cite-74}, and Manga109 \cite{cite-75}. The results, detailed in Table \ref{tab:nature}, clearly attest to our model's preeminence over a spectrum of state-of-the-art methods. Our approach is particularly noteworthy for its exceptional performance relative to ESRGAN, highlighting the efficacy of integrating multimodal data to bolster feature extraction capabilities. This enhancement allows for more nuanced texture details to be discerned from the images, as evidenced by the results.

\begin{figure}[htbp]
	\begin{minipage}[t]{1.0\linewidth}
		\centering
		\includegraphics[width=0.75\linewidth]{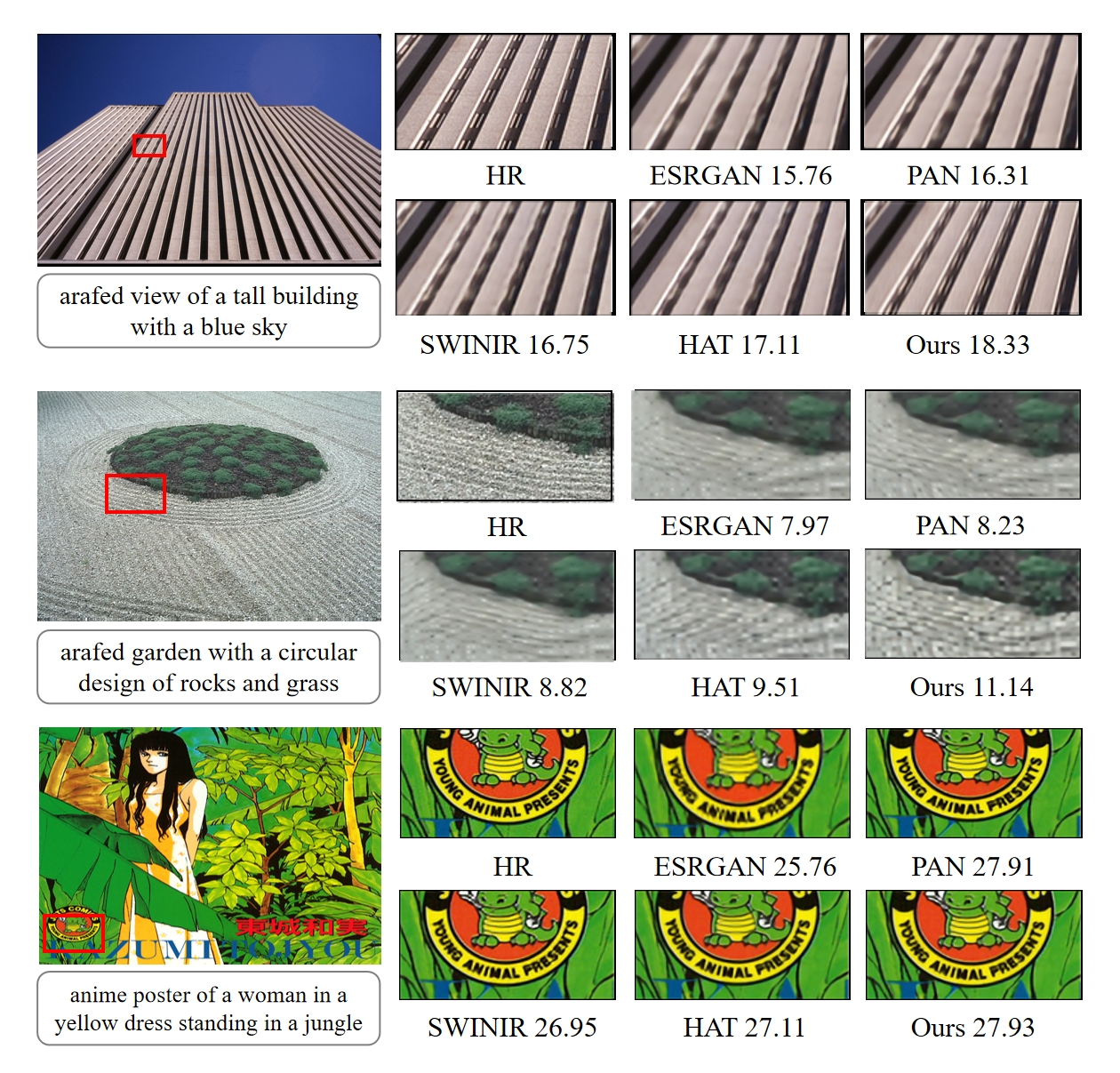}
	\end{minipage}
\caption{Visualization of enhanced image details for select test images, accompanied by captions generated by the semantic extraction module. The displayed PSNR values quantify the image quality.}	
\label{fig:visual}
\end{figure}

In Figure \ref{fig:visual}, we present a detailed visual comparison that showcases our model's adeptness at image detail restoration. The visualizations vividly convey the model's advanced ability to reconstruct color fidelity and intricate textures, outstripping its peers. This superior performance can be attributed to the model's augmented representational capacity, derived from the integration of content-aware attention mechanisms. These mechanisms enable the model to focus selectively on the most pertinent features within the image, thereby preserving vital details with exceptional accuracy.

Furthermore, our analysis indicates that our model maintains its high performance even when tasked with processing images of small dimensions. This robustness is likely due to the model's capacity to leverage semantic information within the image, prompting it to generate and reconstruct richer content. The result is a more authentic and detailed representation of the image's nuances.

\subsubsection{Enhanced Performance on Document Image Super-Resolution}

\begin{table}[htb]
 \centering
  \caption{Comparative Performance on Document Image Super-Resolution}
  \label{tab:document}
  \resizebox{0.55\textwidth}{!}{%
  \begin{tabular}{lccc}
    \toprule
    Method & Scale & PSNR (dB) & SSIM \\
    \midrule
    ESRGAN & $\times$2 & 25.53 & 0.9441 \\
    PAN & $\times$2 & 25.97 & 0.9489 \\
    RealSR & $\times$2 & 26.64 & 0.9563 \\
    SwinIR & $\times$2 & 29.13 & 0.9691 \\
    HAT & $\times$2 & 29.89 & 0.9743 \\
    Ours & $\times$2 & \textbf{31.05} & \textbf{0.9773} \\
    \midrule
    ESRGAN & $\times$3 & 23.97 & 0.9382 \\
    PAN & $\times$3 & 24.92 & 0.9411 \\
    RealSR & $\times$3 & 25.30 & 0.9528 \\
    SwinIR & $\times$3 & 27.61 & 0.9603 \\
    HAT & $\times$3 & 27.92 & 0.9627 \\
    Ours & $\times$3 & \textbf{28.79} & \textbf{0.9687} \\
    \midrule
    ESRGAN & $\times$4 & 23.11 & 0.9361 \\
    PAN & $\times$4 & 23.61 & 0.9437 \\
    RealSR & $\times$4 & 23.59 & 0.9432 \\
    SwinIR & $\times$4 & 24.90 & 0.9427 \\
    HAT & $\times$4 & 25.08 & 0.9469 \\
    Ours & $\times$4 & \textbf{25.41} & \textbf{0.9503} \\
    \bottomrule
  \end{tabular}
  }
\end{table}

A rigorous set of experiments was performed using the BDDI dataset \cite{cite-66}, and the results are systematically detailed in Table \ref{tab:document}. The data clearly demonstrates that our proposed model achieves superior performance metrics, notably in Peak Signal-to-Noise Ratio (PSNR) and Structural Similarity Index Measure (SSIM), surpassing other contemporary models.

\begin{figure}[htbp]
  \begin{minipage}[t]{1.0\linewidth}
    \centering
    \includegraphics[width=0.75\linewidth]{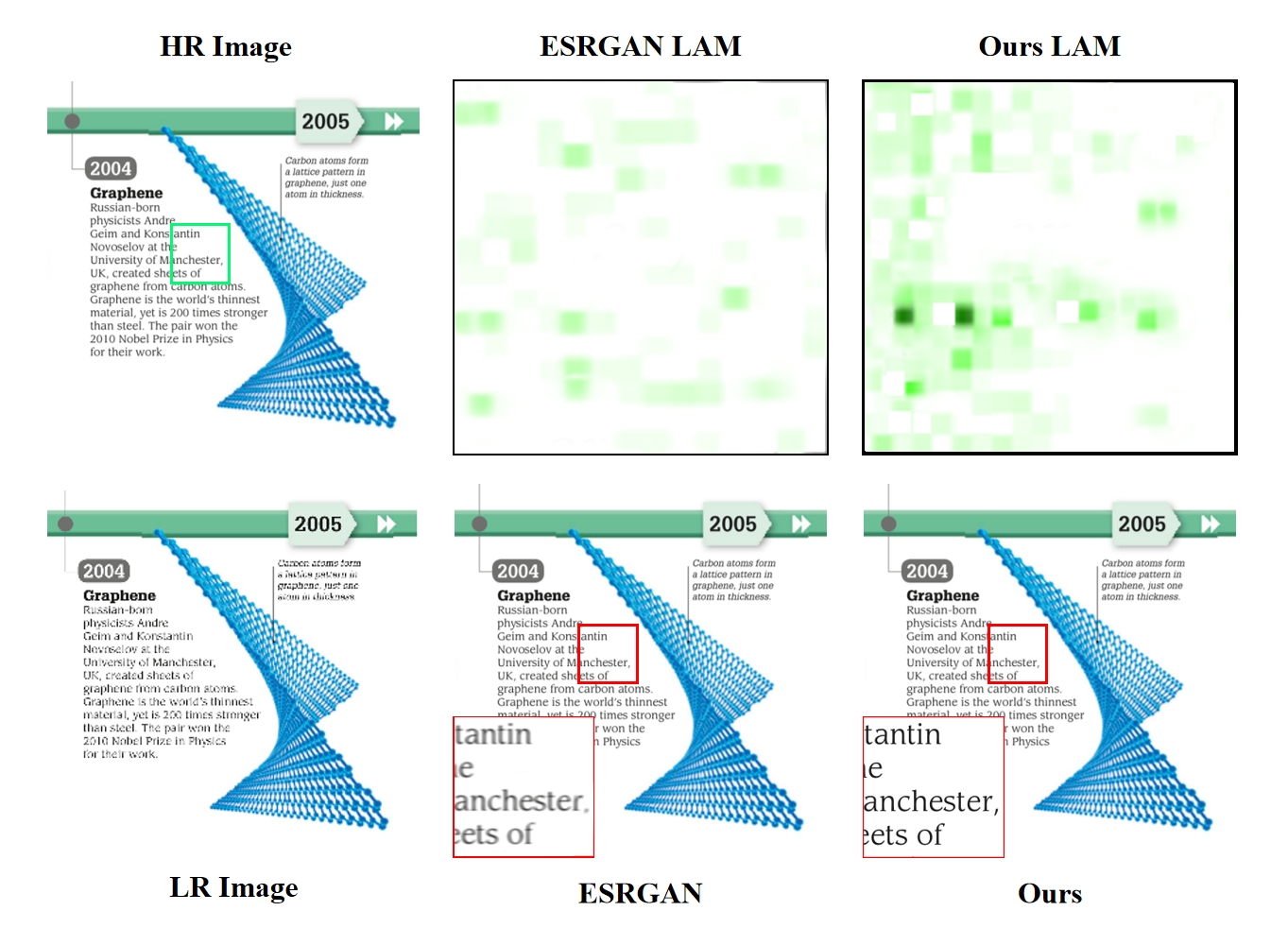}
  \end{minipage}
\caption{Visualization and analysis of the Locally Adaptive Modulation (LAM) effect on text detail enhancement in the BDDI dataset. Our approach significantly improves pixel activation in text-heavy areas.}  
\label{fig:lam}
\end{figure}

Additionally, Figure \ref{fig:lam} illustrates the enhanced edge recovery of intricate text details achieved by our method. An in-depth investigation into the Locally Adaptive Modulation (LAM) characteristic \cite{cite-70} reveals that our model not only enhances text region activation but also adeptly manages pixel activation in adjacent areas, thus sharpening the focus on salient content within the images.

It is noteworthy that the dataset contains instances of text blurring due to small image dimensions, which present challenges in model training. Nevertheless, our findings indicate that the application of a multimodal training approach yields images with significantly improved clarity and sharpness in text regions, as opposed to training without it. This enhancement is likely attributable to the model's deeper comprehension of image content, facilitated by leveraging pre-trained models from the NLP domain within the multimodal framework. Consequently, our approach is capable of producing superior results even when the original images are blurred.

\subsection{Ablation Studies}

\subsubsection{Effectiveness of Each Component}

\begin{table}[htb]
\centering
  \caption{Ablation study results. $\oslash PE$ denotes the exclusion of the position encoding module, $\oslash CMA$ represents the substitution of the content mask attention, and $\oslash MT$ signifies the absence of multimodal training.}
  \label{tab:ablation}
  \resizebox{0.55\textwidth}{!}{%
  \begin{tabular}{lccc}
    \toprule
					Method & Scale & BSD100 & BDDI \\
					\midrule
					Baseline    & $\times$2 & 32.10 / 0.9024 & 31.05 / 0.9773 \\
					  $\oslash PE$ & $\times$2 & 31.91 / 0.8982 & 28.71 / 0.9682 \\
					$\oslash CMA$ & $\times$2 & 31.69 / 0.8936 & 28.03 / 0.9679 \\
					$\oslash MT$ & $\times$2 & 31.62 / 0.8902 & 25.53 / 0.9441 \\
					\midrule
					Baseline    & $\times$3 & 29.71 / 0.8615 & 28.79 / 0.9687 \\
					$\oslash PE$ & $\times$3 & 29.58 / 0.8610 & 27.16 / 0.9604 \\
					$\oslash CMA$ & $\times$3 & 29.63 / 0.8613 & 26.67 / 0.9553 \\
					$\oslash MT$ & $\times$3 & 29.44 / 0.8607 & 23.97 / 0.9382 \\
					\midrule
					Baseline    & $\times$4 & 27.76 / 0.7481 & 25.41 / 0.9503 \\
					$\oslash PE$ & $\times$4 & 27.71 / 0.7467 & 25.13 / 0.9479 \\
					$\oslash CMA$ & $\times$4 & 27.66 / 0.7459 & 24.89 / 0.9446 \\
					$\oslash MT$ & $\times$4 & 27.64 / 0.7406 & 23.11 / 0.9361 \\
					\bottomrule
\end{tabular}
}
\end{table}

To thoroughly examine the contribution of individual components to our methodology's performance, we executed a series of ablation studies, the outcomes of which are presented in Table \ref{tab:ablation}. These investigations scrutinized three pivotal aspects of our framework, with performance evaluations conducted on both the BSD100 and BDDI datasets.

The exclusion of the position encoding module ($\oslash PE$) led to a discernible decrease in performance metrics across both datasets, with a more marked impact on document images. This observation suggests that the positional cues provided by the encoding module are especially critical for accurately resolving smaller text regions within images. The replacement of the content mask attention ($\oslash CMA$) with a straightforward feature summation also resulted in diminished performance metrics, reinforcing the efficacy of our content-focused multimodal feature fusion strategy in enhancing image super-resolution tasks. The removal of the semantic extraction module ($\oslash MT$), which effectively reduced our approach to a standard image super-resolution model, caused a substantial drop in performance metrics. This highlights the value of leveraging techniques from diverse domains through multimodal training to significantly improve outcomes, even for low-level vision tasks. Such insights may have far-reaching implications for a variety of applications, including image denoising, as noted by recent studies \cite{newcite-24}.

The findings from these ablation studies provide an in-depth understanding of the importance of each component in our approach and emphasize the advantages of adopting a multimodal strategy to advance the capabilities of various computer vision tasks.

\subsubsection{Exploring the Effects of Content Area Mask Magnitude on Model Performance}

\begin{table}[htb]
\centering
\caption{Comparative analysis of model performance with varying content area mask magnitudes. The data indicates that a moderate increase in magnitude can enhance model efficacy.}
\label{tab:mask}
\resizebox{0.7\textwidth}{!}{%
\begin{tabular}{lccc}
\hline
Magnitude & Upscaling Factor & BSD100         & BDDI           \\ \hline
0.0       & $\times$2        & 31.94 / 0.9003 & 30.62 / 0.9711 \\
\textbf{0.3}       & $\times$2        & 32.10 / 0.9024 & 31.05 / 0.9773 \\
0.5       & $\times$2        & 31.54 / 0.8895 & 30.22 / 0.9696 \\
0.7       & $\times$2        & 31.88 / 0.8936 & 29.32 / 0.9603 \\
1.0       & $\times$2        & 31.47 / 0.8872 & 26.33 / 0.9481 \\ \hline
\end{tabular}%
}
\end{table}

This section delves into the effect of varying the magnitude of content area masks on the efficacy of our model. The content area mask is a pivotal element in our experiments and is quantified by distinct magnitudes. Table \ref{tab:mask} encapsulates the outcomes of our comparative analysis, wherein we experimented with magnitudes ranging from 0.0 to 1.0, while maintaining a consistent upscaling factor of $\times$2. The model's performance was appraised using two separate datasets.

A key observation is that a magnitude of 0.3 yields the most substantial enhancement in model performance, as evidenced by the superior scores in image quality and associated metrics. This suggests that an optimal, moderate increment in the magnitude of the content area mask, exemplified by a value of 0.3, can significantly bolster the capabilities of our model. The results underscore the importance of fine-tuning the magnitude of the content area mask to refine and augment the performance of our model. Additional experimentation with this variable may yield further insights, potentially contributing to advancements in image processing and restoration methodologies.

\subsubsection{Influence of Loss Weighting on Multimodal Performance}

\begin{table}[htb]
\centering
\caption{Comparative experiment outcomes based on varying loss weights. "Weight" refers to the semantic similarity loss weight, with pixel loss and visual feature similarity loss weights fixed at 1.0 and 0.0001, respectively.}
\label{tab:weight}
\resizebox{0.6\textwidth}{!}{%
\begin{tabular}{lccc}
\hline
Weight         & Scale                         & BSD100                             & BDDI                               \\ \hline
0.0            & $\times$2 & 31.25 / 0.8879 & 25.73 / 0.9467 \\
1.0            & $\times$2                     & 26.13 / 0.8122                     & 21.39 / 0.8951                     \\
0.1            & $\times$2                     & 29.37 / 0.8890                     & 24.98 / 0.9381                     \\
0.01           & $\times$2                     & 31.68 / 0.8883                     & 29.11 / 0.9695                     \\
\textbf{0.001} & $\times$2                     & \textbf{32.10 / 0.9024}                     & \textbf{31.05 / 0.9773}                     \\
0.0001         & $\times$2                     & 31.89 / 0.8991                     & 30.04 / 0.9711                     \\ \hline
\end{tabular}%
}
\end{table}

Our investigation extends to the influence of loss weights on our model's multimodal performance. Loss weights represent the relative significance assigned to the different components of loss during the training of our model. We specifically examine the impact of varying the semantic similarity loss weight, while maintaining the pixel loss and visual feature similarity loss weights at empirical values of 1.0 and 0.0001, respectively.

The summary of outcomes from our comparative experiments is presented in Table \ref{tab:weight}. We experimented with loss weight values ranging from 0.0 to 0.0001, while the upscaling factor remained constant at $\times$2. The model's performance was evaluated on two distinct datasets, BSD100 and BDDI, focusing on the quality of scale enhancement. The empirical data indicates that the choice of semantic similarity loss weight markedly influences the model's efficacy. A decline in the weight of this loss component correlates with enhanced upscaling results, with an optimal weight of 0.001 yielding the most favorable performance. This underscores the importance of a calibrated approach to loss component weighting in our model to achieve optimal results. It should be emphasized that the ideal loss weight configuration may vary across different datasets and tasks. Therefore, fine-tuning the loss weights is an essential step towards enhancing performance across various modalities.

\section{Conclusion}

This study presents a novel approach to image super-resolution that effectively capitalizes on the semantic content within images to augment the super-resolution process. We have developed a dedicated semantic extraction and processing module, complemented by a sophisticated cross-modal feature fusion module, which is adeptly customized for various image categories. The integration of positional encoding allows for the establishment of precise, localized connections between the extracted semantic elements and their corresponding regions in the image. Our comprehensive experimental analysis validates the superiority of our method, as evidenced by significant improvements in performance metrics and the quality of the visual results. This research makes a substantial contribution to the domain of image super-resolution by introducing an innovative technique that exploits the intrinsic semantics of images to refine model training and enhance the precision of image detail recovery.

\bibliographystyle{elsarticle-num}
\bibliography{references}

\end{document}